\documentclass[preprint,12pt]{elsarticle}

\usepackage{amssymb}
\usepackage{float}
\usepackage{graphicx} 
\usepackage{amsmath}
\usepackage{multirow}
\usepackage{booktabs}
\usepackage{comment}
\usepackage{url}
\usepackage{subcaption}
\usepackage{longtable}
\usepackage{xurl}
\usepackage{hyperref}

\journal{Computer Methods and Programs in Biomedicine}

\begin{document}

\begin{frontmatter}

\title{RecallRisk-BERT: Joint Prediction of Severity and Reported Root Cause in FDA Medical Device Recall Records}

\author[ankara]{Ali Semih Atalay}
\author[ankara]{Sevgi Yigit-Sert}

\affiliation[ankara]{
    organization={Department of Computer Engineering, Faculty of Engineering, Ankara University},
    city={Ankara},
    postcode={06830},
    country={Türkiye}
}

\begin{abstract}
\textbf{Background and objectives:}
Recall severity and reported root cause provide complementary information
for reviewing medical device recalls, yet previous studies have modeled them
separately. We examined whether these two outcomes could be classified
jointly using 54,165 FDA medical device recall records from openFDA. 

\noindent \textbf{Methods:}
RecallRisk-BERT combines a fine-tuned
PubMedBERT encoder for recall narratives with embeddings of product code,
regulation number, and medical specialty to predict recall severity (Classes
I–III) and reported root cause (nine consolidated categories). We evaluated
all models on a stratified random split, with paired bootstrap confidence
intervals for selected comparisons, and three models on a chronological
split.

\noindent \textbf{Results:}
Under the random split, LightGBM achieved the highest severity 
macro-F1 (0.856), as well as the highest root-cause
macro-F1 (0.798). RecallRisk-BERT's severity macro-F1 (0.852) did not differ
significantly from LightGBM's and exceeded that of single-task PubMedBERT
(0.806), whereas its root-cause macro-F1 was lower (0.710 vs. 0.726).
RecallRisk-BERT also identified more Class I records than LightGBM (89.9\% vs.
84.3\%; 
95\% CI, 3.9–7.3). Observed and
model-derived mean severity scores ranked the nine root-cause categories
similarly (Spearman's $\rho$ = 0.983). On later records, all three models scored
lower; single-task PubMedBERT achieved the highest chronological macro-F1 for
severity (0.494) and root cause (0.292), followed by RecallRisk-BERT (0.458;
0.278) and LightGBM (0.358; 0.184). 

\noindent \textbf{Conclusions:} 
Fine-tuned text–tabular transformer models scored higher than
gradient boosting on later recall records, whereas joint learning improved
severity only under the random split. Evaluation on reports as first
submitted is needed before using these models for post-report triage.

\end{abstract}

\begin{keyword}
Medical device recall \sep Multi-task learning \sep Text–tabular fusion \sep PubMedBERT \sep Chronological validation\sep FDA
\end{keyword}

\end{frontmatter}

\section{Introduction}
\label{sec:intro}

Medical devices have become an integral and vital part of healthcare services.
They encompass a broad variety of products, from simple items such as medical gloves and bandages to complex machines like blood pressure monitors and X-ray devices.
The development and production of medical devices is a complex and demanding process that requires high-quality standards and ensures that the final product complies with existing regulatory requirements. 
This process has a multi-stage and interdisciplinary structure, extending from conceptual design to a fully functional product ready for clinical use~\cite{ocampo2019}. At any of these stages—such as design, manufacturing, or software development—deviations in quality and safety may lead to medical device recall cases.

A recall is the process of removing or correcting products that violate standards and is overseen by the U.S. Food and Drug Administration (FDA). Recalls are voluntary actions taken by manufacturers and distributors to protect public health and welfare by withdrawing products that pose significant risks, such as injury, or are otherwise defective. The protocol known as 21 CFR Part 7 provides guidance to responsible companies for conducting effective recalls. However, in some cases, if a manufacturer or importer does not voluntarily recall a device that poses a health risk, the FDA can issue a mandatory recall order under 21 CFR Part 810 (Medical Device Recall Authority). 

Hundreds of medical device recalls occur each year~\cite{Gagliardi2017}. It has been reported that, in 2022, FDA recorded 70 high-risk recalls, compared to the five-year average of 47~\cite{Taylor2023}. 
Medical device recalls lead to significant financial losses for manufacturers~\cite{Marucheck2011}, reputational damage~\cite{Blom2022}, and disruptions in the supply chain~\cite{Ahsan2014, Mukherjee2018, hu2026}. At the same time, they pose direct clinical risks to patients and healthcare professionals. 
Devices regulated by the FDA are classified into three categories based on the level of risk they pose to patients: Class I (low risk), Class II (moderate risk), and Class III (high risk). 
Medical device regulatory classes and recall severity classes use different risk logics. While Class III devices generally indicate higher regulatory device risk, Class I recalls involve situations where exposure to a violative product may reasonably cause serious health problems or death. Class II recalls involve products that may cause temporary or medically reversible adverse health consequences, and Class III recalls involve products unlikely to cause adverse health consequences~\cite{FDA2024}. 
Life-supporting devices, such as implantable cardioverter defibrillators, are categorized as Class III devices; however, when a recall of such devices poses the highest level of patient risk, it is classified as a Class I recall~\cite{Sarkissian2018, Villarraga2007}.
Since this classification directly reflects the potential clinical impact of a recall, it is critically important not only to model whether a recall will occur but also to accurately model the severity of the recall.

The increasing prevalence of recalls has triggered growing interest in proactive prediction models within the literature. A significant portion of the studies on medical device recalls has focused on binary classification problems aimed at predicting whether a recall will occur or not. More recent studies have proposed approaches for predicting recall causes (root-cause or recall initiator) as multi-class classification problems. However, existing studies generally address these two problems independently and do not jointly model the relationship between recall severity and the underlying error mechanism. Yet, it is known that different error mechanisms produce different levels of clinical risk. The FDA’s recall classification process itself evaluates factors such as the nature and potential health risk of the harm, the scope and cause of the defect, the probability of occurrence, and the level of danger posed to patients together~\cite{FDA2024}.  
A study on Class I recalls shows that design defects, software-related failures, and sterility issues are among the major root-causes associated with high-risk recalls, while labeling and packaging problems are also among the frequently reported recall causes~\cite{mj2024}. 

To the best of our knowledge, an integrated (joint) prediction framework that models these two targets together is still absent in the literature. 
Moreover, the structural characteristics of FDA recall databases create significant class imbalance problems from a modeling perspective. While Class I recalls occur less frequently compared to Class II recalls, certain root-cause categories — especially software design in AI/ML devices — appear to be highly concentrated ~\cite{Chen2025, Gao2025}. Such imbalanced distributions may create bias toward majority classes, particularly in single-task classification models, and may make it more difficult to distinguish minority classes. The accurate identification of high-risk recalls, which are especially critical for patient safety, makes this problem even more important.

Within the scope of this study, the following research questions were investigated:
\begin{itemize}
    \item RQ1: How accurately can recall severity be predicted from recall narratives and structured
    device/regulatory information?
    \item  RQ2: How does joint learning affect severity and root-cause prediction compared with equivalent
    single-task models?

    \item RQ3: How do observed severity distributions differ across root-cause categories, and to what
    extent are these patterns reflected in model predictions?
    \item RQ4: How well do models trained on earlier records generalize to later
  records?

\end{itemize}

To answer these research questions, a multi-task learning framework that jointly learns recall severity (recall class) and recall cause (root-cause category) is proposed. In the proposed approach, recall texts (reason for recall and product description) are represented using a transformer-based model pretrained on biomedical texts (e.g., PubMedBERT). In addition, these features are combined with structured (tabular) data such as product code, regulation number, and medical specialty area to construct a rich feature space.

The main contributions of this study can be summarized as follows:
\begin{itemize}
    \item The predictability of recall severity is systematically investigated on 54,165 FDA recall records using a broad set of machine learning and deep learning models.
    \item A multi-task learning framework, RecallRisk-BERT, is proposed to jointly model recall severity and root-cause category.
    \item Severity patterns across reported root-cause categories are investigated through both statistical (ground-truth) and model-based analyses.
    \item Three main modeling approaches are compared under random and chronological evaluation, showing that the top-ranked model changes on later records and that fine-tuned transformer models decline less than gradient boosting.
    \item Class I recall and precision are reported separately, revealing a trade-off that is not apparent from overall macro-F1 alone.
\end{itemize}

The aim of this study is not to predict whether a medical device will be recalled, since all observations correspond to recalls that have already occurred. Instead, the study investigates joint prediction of recall severity and root-cause category from recall narratives and structured device/regulatory information. This task is motivated by the need to efficiently review and prioritize recall reports,
for which severity and root-cause information provide complementary perspectives. 
Our study evaluates this approach retrospectively using
publicly available FDA recall records. 

The rest of the paper is organized as follows: Section~\ref{sec:relatedWork} reviews related work on recall prediction; Section~\ref{sec:method} describes the dataset and methodology; Section~\ref{sec:expSetup} explains the experimental setup; and Section~\ref{sec:results} presents the results, including chronological evaluation and Class I analysis; Section~\ref{sec:discussion} discusses the findings and limitations; and Section~\ref{sec:conc} concludes the paper.

\section{Related Work}
\label{sec:relatedWork}

This section first summarizes computational work on post-market surveillance of medical devices, then reviews the literature on prediction and analysis of medical device recalls. The last subsection compiles the key gaps in the existing literature and justifies which gaps this study focuses on.


Vlachos et al.~\cite{vlachos2003} developed an electronic post-market surveillance system to collect and share reports of adverse incidents involving medical devices. More recently, Chen et al.~\cite{Chen2025} analyzed recalls of AI/ML-enabled medical devices from 1997 to 2024 and
reported that software design flaws accounted for a substantial share of these recalls. Therefore, analyzing and predicting medical device failures and recalls has become a critical research area for both manufacturers and regulatory bodies.

\subsection{Prediction and analysis of medical device recalls}

\noindent \textit{Descriptive and Epidemiological Studies.} Villarraga et al.~\cite{Villarraga2007} established the basic conceptual framework of the field by presenting a descriptive analysis of FDA recalls. Thirumalai and Sinha \cite{Thirumalai2011} empirically examined the sources of recalls in the medical device industry and their financial consequences, reporting that device design, manufacturing processes, and labeling errors were the primary triggers. More recently, Sarkissian \cite{Sarkissian2018} analyzed 871 Class I recalls from 2014–2018, reporting that the dominant root-causes were ``packaging'' (47.4\%), ``component'' (14\%), and ``design'' (13.3\%). M.J. et al.~\cite{mj2024} examined Class I recalls from 2020–2023 and reported that injury risks, erroneous results, software problems, and manufacturing defects are the dominant root-cause categories. While these studies offer valuable descriptive findings, they have remained largely descriptive and have not provided opportunities for proactive prediction.

\noindent \textit{Recall Prediction and Classification Approaches.} The use of machine learning methods on medical device recall data has become increasingly common in recent years. Emakhu et al.~\cite{emakhu2019} developed a machine learning framework to predict recall failure types using software-based medical device recall records published by the FDA; approximately 84\% accuracy was achieved with the multilayer perceptron (MLP) model.
Slivinskis et al.~\cite{Barbosa2025} presented a random forest-based algorithm using data from open sources such as PubMed and Google Trends; they demonstrated that they could predict recall status with high accuracy 3, 6, and 12 months in advance. The strength of the study is that it provides evidence of the predictive power of signals from external (non-FDA) data sources; its limitation is its small sample size and the use of external media/literature signals rather than internal device features.
Zhu et al.~\cite{zhu2025} proposed a deep learning-based prediction structure on a dataset covering 45,398 devices validated via 510(k) paths between 2003 and 2020. Their method models the structural features of the predicate citation network using graph learning methods and captures the temporal patterns of citation network features. The results showed that the network-based approach provided significant improvements in both accuracy and timeliness compared to classical tabular methods. Everhart et al.~\cite{everhart2023} found statistically significant relationships between the predictive properties of 510(k) devices and subsequent recalls using linear probability models. Zhalechian et al.~\cite{Zhalechian2024} developed a gradient-boosting-based human–algorithm framework to estimate future recall risk for more than 31,000 FDA 510(k) devices and support clearance decisions under FDA workload constraints.

\noindent \textit{Root-Cause Estimation Approaches.} Hu et al.~\cite{hu2024} employed a machine learning–natural language processing workflow to analyze medical device recall causes, using DBSCAN clustering to identify recall initiator categories and text-similarity-based classification to support practitioner-oriented insights. 
In their later work, Hu et al.~\cite{hu2026} argued that the medical device recall problem should be addressed from a supply chain risk management and resilience perspective and developed a recall initiator prediction framework that combines optimized feature selection and clustering techniques to strengthen error detection at the premarket stage. Highlighting the contribution of practical features such as ``product code'' to predictive power, the study achieved an accuracy level of 88.85\% for high-risk Class II and Class III devices.

Table~\ref{tab:literature_comparison} provides a comparative overview of the methodologies of existing studies.

\begin{table}[!ht]
\vspace{0.3cm}
\caption{Comparison of methodological characteristics of major studies in the medical device recall prediction literature.}
\label{tab:literature_comparison}
\footnotesize
\setlength{\tabcolsep}{4pt}
\renewcommand{\arraystretch}{1.25}
\begin{tabular}{@{}p{3cm}p{2.5cm}p{2.5cm}p{3cm}p{2.8cm}@{}}
\toprule
\textbf{Study} & \textbf{Prediction Target} & \textbf{Task Type} & \textbf{Methodology} & \textbf{Input Modalities} \\
\midrule
Sarkissian \cite{Sarkissian2018} & Root-cause distribution & Descriptive & Statistical analysis & Tabular only \\
Emakhu et al.\ \cite{emakhu2019} & Failure type & Multiclass classification & MLP + ML & Tabular only \\
Everhart et al.\ \cite{everhart2023} & Recall occurrence & Binary Prediction & Linear probability & Tabular only \\
Hu et al.\ \cite{hu2024} & Recall initiator & Unsupervised & DBSCAN + NLP & Text only \\
M.J. \ et al.\ \cite{mj2024} & Root-cause distribution & Descriptive & Statistical analysis & Tabular only \\
Zhalechian et al.\ \cite{Zhalechian2024} & Recall occurrence & Binary classification & Gradient boosting & Tabular only \\
Slivinskis et al.\ \cite{Barbosa2025} & Recall occurrence & Binary classification & Random forest & Tabular + external signals (PubMed/Trends) \\
Chen et al.\ \cite{Chen2025} & Root-cause distribution & Descriptive & Statistical analysis & Tabular only \\
Hu et al.\ \cite{hu2026} & Recall initiator & Multiclass & Random forest + DBSCAN & Tabular only \\
Zhu et al.\ \cite{zhu2025} & Recall occurrence & Binary classification & Graph + temporal DL & Tabular + network structure \\
\midrule
\textbf{This study} & \textbf{Recall severity + root-cause} & \textbf{Multi-task multiclass} & \textbf{PubMedBERT + tabular fusion + MTL} & \textbf{Text + tabular (multimodal)} \\
\bottomrule
\end{tabular}

\smallskip
\footnotesize\textit{Note:} DL: Deep Learning; MTL: Multi-Task Learning; MLP: Multi-Layer Perceptron; DBSCAN: Density-Based Spatial Clustering of Applications with Noise.
\end{table}

\subsection{Research Gap}
This study is designed to explicitly address the following gaps in the medical device recall estimation literature.

\noindent \textit{Gap 1 — The severity level is not directly modeled as a multi-class target.} The vast majority of studies either treat the recall event as a binary (recall/no-recall) target or focus only on Class I recalls. Zhalechian et al.~\cite{Zhalechian2024} optimized their models for a binary target, then reported individual AUC values post-hoc according to severity classes; however, their models were not structurally designed for a multi-class severity target. 
None of the studies in Table~\ref{tab:literature_comparison} treats recall severity (Class I, II, and III) as a multi-class prediction target.


\noindent \textit{Gap 2 — The severity and root-cause dimensions are treated separately.} Current approaches either target only the recall event or only the root-cause. 
To the best of our knowledge, the multi-task learning (MTL) paradigm has not yet been adapted to the context of medical device recall prediction. This study addresses this gap by proposing an MTL framework that simultaneously predicts severity and root-cause through shared representations. 

\noindent \textit{Gap 3 — Insufficient integration of textual and tabular data.} Although FDA recall records contain rich textual descriptions (reason for recall, product descriptions) and regulatory tabular fields (product code, regulation number, specialty area), current prediction studies often treat these two data types separately: some use only tabular features, while others model only textual information. Although the success of biomedical domain-specific language models (PubMedBERT \cite{gu2021}, BioBERT \cite{lee2020}, ClinicalBERT \cite{huang2019}) in health text classification has been demonstrated in various studies \cite{li2024, luschi2023}, the application of these models to direct recall severity estimation and their use within a multimodal architecture with tabular features has not been systematically addressed in the literature. This study responds to this gap by adopting a multimodal learning approach that combines biomedical BERT representations with tabular feature embeddings.

\section{Materials and Methods}
\label{sec:method}

\subsection{Data source}

In this study, FDA medical device recall records covering the period from November 2002 to October 2025 were obtained in JSON format from the openFDA platform\footnote{https://open.fda.gov/apis/device/recall/download/ (accessed December 2025)}. After removing duplicate records and records with missing data, the final dataset comprised 54,165 recall records. The unit of analysis was a product-level recall record, identified by the
``product res number''. A single recall event can comprise multiple such records.
We included all medical device categories available in the FDA medical device recall database, without imposing any restriction in terms of medical specialty area or approval pathway. 
From the raw openFDA records, 11 relevant fields (as shown in Table~\ref{tab:recall_dataset_fields}) were retained for dataset construction, target definition, preprocessing, and model development. The primary inputs of the model consisted of two textual fields (reason for recall and product description) and three structured categorical fields (product code, regulation number, and medical specialty description). The remaining fields were used for target construction, descriptive analysis, or the chronological split, as described in the following sections.
In particular, recall class was used as the primary target variable, while root-cause description was used only to derive the consolidated root-cause category target and was not included as an input feature.

\begin{figure}[!t]
\centering
\begin{subfigure}[b]{0.45\textwidth}
    \centering
    \includegraphics[width=\textwidth]{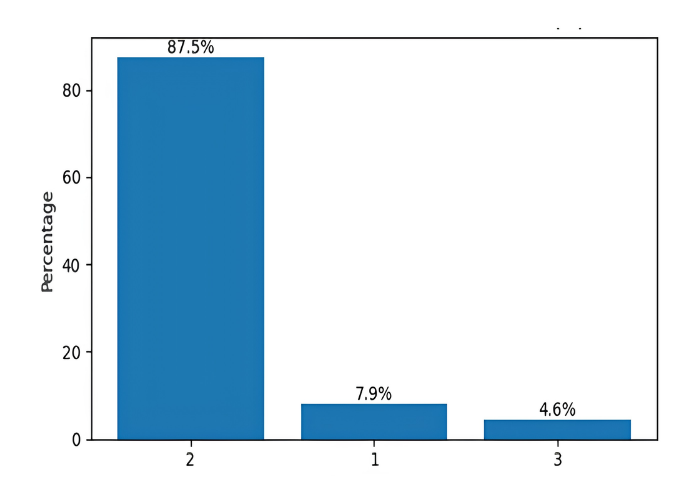}
    \caption{}
    \label{fig:sub1}
\end{subfigure}
\hfill
\begin{subfigure}[b]{0.50\textwidth}
    \centering
    \includegraphics[width=\textwidth]{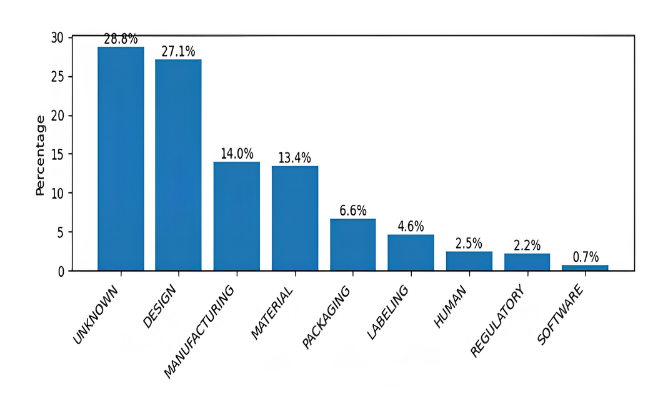}
    \caption{}
    \label{fig:sub2}
\end{subfigure}
\vspace{-0.1cm}
\caption{Class distributions of (a) recall class and (b) root-cause category. The numeric labels correspond to Class I, Class II, and Class III recall categories.}
\label{fig:distribution}
\end{figure}

The dataset has a significant class imbalance in both recall severity and root-cause dimensions. In the recall class variable, Class II constitutes the dominant majority with 87.5\%, while Class I is represented at 7.9\% and Class III at 4.6\%. The distribution of the secondary target variable, root-cause category, is: Unknown (28.8\%), Design (27.1\%), Manufacturing (14.0\%), Material (13.4\%), Packaging (6.6\%), Labeling (4.6\%), Human (2.5\%), Regulatory (2.2\%), and Software (0.7\%). The distribution of target classes is shown in Figure~\ref{fig:distribution}.

\begin{table}[!t]
\centering
\footnotesize
\caption{Selected openFDA fields used for dataset construction and modeling.}
\vspace{0.2cm}
\label{tab:recall_dataset_fields}
\begin{tabular}{ p{2cm}  p{7cm} l}
\hline
\textbf{Field} & \textbf{Description} & \textbf{Type} \\
\hline

product res number & Internal recall identifier for each recalled device. & Identifier \\
\hline
product description & Brief description of the recalled product. & String \\
\hline
product code & Three-letter code identifying the device category, based on its 21 CFR classification, technology, and intended use. & Categorical \\
\hline
recalling firm & Firm initiating the recall or primarily responsible for the product's manufacture/marketing. & Categorical \\
\hline
reason for recall & Description of the product defect and how it violates the FD\&C Act. & String \\
\hline
root-cause description & FDA-determined general type of recall cause. & Categorical \\
\hline
device name & Proprietary or trade name of the cleared device. & Categorical \\
\hline
medical specialty description & Medical specialty assigned by the device regulation. & Categorical \\
\hline
regulation number & CFR regulation under which the device is classified. & Categorical \\
\hline
recall class & Recall classification based on the reason for recall. & Integer \\
\hline
event date initiated & Date the firm first notified the public or consignees. & Date \\
\hline

\end{tabular}
\end{table}

\subsection{Target variable construction}
\label{sec:3.2}
The primary target was the FDA-assigned \textit{recall class}
(Class~I, Class~II, or Class~III; see Table~\ref{tab:recall_dataset_fields}). Class I recalls represent the highest-risk recalls, carrying a probability of serious health risk or death; Class II recalls represent moderate-risk recalls that may lead to temporary or medically reversible health problems; and Class III recalls represent the lowest-risk recalls, which are not expected to pose a serious health risk.
The secondary target variable, \textit{root-cause category},
represents the reported underlying cause of the recall and was
derived from the root-cause description field in the FDA records. The 43 original
labels were consolidated into nine study-defined categories: Unknown,
Design, Manufacturing, Material, Software, Labeling, Packaging, Human,
and Regulatory. The complete mapping is provided in Supplementary
Table~S1.
Design includes software, package, and labeling design-related causes,
whereas Software, Packaging, and Labeling contain the other causes
assigned to these groups. Unknown comprises ``Other'',
``Under Investigation by firm'', ``Unknown/Undetermined by firm'',
and ``Pending''; labels were not assigned to this category solely
because they were infrequent.

To obtain a continuous model-derived severity estimate, the predicted class probabilities from the recall severity head were converted into an Expected Recall Severity Score. Since FDA Class I recalls represent the highest risk level and Class III recalls represent the lowest risk level, the class probabilities were weighted as follows:

\vspace{-0.02cm}
\begin{equation}
\resizebox{0.9\textwidth}{!}{$
    \text{Expected Recall Severity Score} = 3 P(\text{Class I}) + 2 P(\text{Class II}) + 1 P(\text{Class III})
    $}
    \label{eq:expectedRisk}
\end{equation}

The score ranges from 1 to 3, where higher values indicate a greater predicted recall severity. The weights represent a chosen ordinal coding rather than measured differences in clinical harm. This score was used only for descriptive model-based analysis, not as an additional training target
or a probability of patient harm. 

\subsection{Feature construction and leakage control}

Model inputs comprised textual and structured features. The textual input was constructed by concatenating the reason for recall and product description fields using a separator token and tokenized with a maximum sequence length of 128 tokens. The structured inputs consisted of
product code, regulation number, and medical specialty description. Each categorical variable was mapped to an integer identifier, with an ``UNK'' token reserved for categories not observed in the
training set. Target labels were numerically encoded. All encodings were fitted on the training set only to avoid preprocessing leakage~\cite{bouke2023}.

To reduce potential shortcut learning, device name and recalling firm were excluded as separate input features. Their exclusion does not necessarily remove device or firm identifiers embedded in the textual fields. The root cause description field was also excluded because it was used to construct the secondary target. Neither target label was included among the model inputs.
Input fields were used as available in the downloaded records. Historical versions were not reconstructed, and the availability of these fields at initial reporting was not verified.

\subsection{Proposed RecallRisk-BERT architecture}

\begin{figure}[!t]  
\centering
\includegraphics[width=\textwidth]{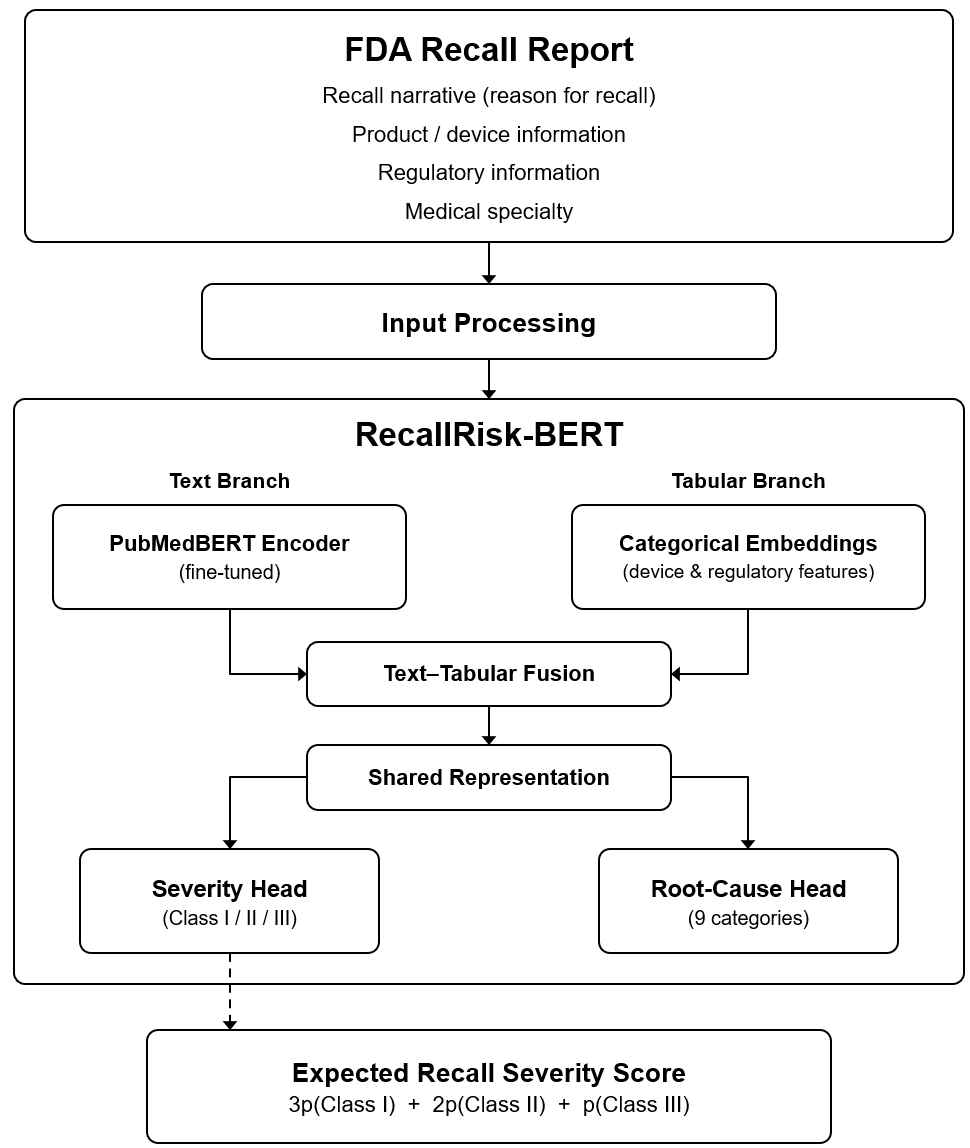}
\caption{Proposed RecallRisk-BERT architecture integrating PubMedBERT-based textual representations and tabular embeddings for joint prediction of recall severity and root-cause category.}
\label{fig:bert_model}
\end{figure}

As shown in Figure~\ref{fig:bert_model}, RecallRisk-BERT combines a transformer-based text encoder and a tabular embedding module through shared representation layers. Two task-specific heads predict recall severity and reported root cause. The severity probabilities are used to compute the Expected Recall Severity Score (Section~\ref{sec:3.2}) for the descriptive analysis in Section~\ref{sec:5.6}.

\noindent \textbf{Text encoder}. The combined text input is encoded with PubMedBERT~\cite{gu2021}, which has been documented to exhibit superior performance in biomedical terminology. The domain-specific terminology density of FDA recall records, such as manufacturing defects, sterility breaches, contamination, and software problems, constitutes the main rationale for this architectural choice. The contextual vector of the [CLS] token is used as the text representation.
PubMedBERT is fine-tuned jointly with the tabular embedding, fusion, and
task-specific output layers. 

\noindent \textbf{Tabular embedding module}. Three categorical variables are passed through separate embedding layers and transformed into learnable dense vectors. After concatenation, they are passed through an MLP layer to obtain a 128-dimensional tabular representation.

\noindent \textbf{Shared representation and task-specific heads}. A 768-dimensional [CLS] output of PubMedBERT is combined with a tabular representation to form an 896-dimensional fused vector; this vector is then passed through a fully connected layer and transformed into a 512-dimensional shared representation. Two parallel classification heads produce logit outputs for recall class (3 classes) and root-cause category (9 classes), respectively.

\subsection{Multi-task learning objective}
The proposed model is based on a hard parameter-sharing approach: the transformer-based text encoder, the tabular fusion component, and the shared representation layers are shared by both tasks; only the final output heads are separated in a task-specific manner. This design enables the structural dependencies between recall severity and root-cause mechanisms to be implicitly learned within a shared latent representation.
The total loss function is defined as a linear combination of task-specific weighted cross-entropy losses:
\begin{equation}
L_{total} = \lambda_{1} \mathcal{L}_{recall\_severity} + \lambda_{2} \mathcal{L}_{root-cause\_category}
\end{equation}

Here, $\mathcal{L}_{\mathrm{recall\_severity}}$ and $\mathcal{L}_{\mathrm{root-cause\_category}}$ denote class-weighted cross-entropy losses. Class weights control the relative contribution of individual classes within each task, whereas $\lambda_1$ and $\lambda_2$ control the relative contributions of the two tasks. Class weights were calculated as

\begin{equation}
\label{eq:weight}
w_i = \frac{N}{K \cdot n_i} 
\end{equation}
where $N$ is the total number of samples, $K$ is the number of classes, and $n_i$ is the number of samples for class $i$.

\section{Experimental Setup}
\label{sec:expSetup}
\noindent \textbf{Baseline Models:} To evaluate the effectiveness of the proposed RecallRisk-BERT model, single-task baseline models from different model families were employed. The first group consists of classical machine learning models trained with TF-IDF-based text representations. In this group, logistic regression~\cite{lavalley2008} and Support Vector Machine (SVM)~\cite{joachims2002} were used as fundamental text classification baselines.
The second group consisted of tree-based ensemble models, including Random Forest~\cite{breiman2001}, XGBoost~\cite{chen2016}, LightGBM~\cite{ke2017}. 
The third group consists of single-task transformer-based models. These models employed a BERT-based text encoder and a tabular fusion structure, but were trained to predict only a single target variable. In this way, a single-task severity prediction model, a single-task root-cause category prediction model, and the proposed multi-task RecallRisk-BERT model were compared. This comparison was used to assess whether the performance difference originates from shared representation learning.

In addition to the classical machine learning models, two neural baseline architectures were implemented to evaluate the effectiveness of non-transformer deep learning approaches, using GloVe-based~\cite{glove2014} text representations combined with tabular features. 
The first baseline is a deep neural network (DNN)~\cite{lecun2015} consisting of four fully connected layers with 1024, 512, 256, and 128 neurons, respectively, each followed by batch normalization and dropout to mitigate overfitting, with ReLU activation in the hidden layers and softmax in the output layer for multi-class classification. 
The second baseline is a BiLSTM model~\cite{97bidirectional} with an attention mechanism. 
The resulting text representation is concatenated with the tabular features and passed through three fully connected layers of 512, 256, and 128 neurons, each followed by batch normalization and dropout, with a softmax output layer. For both the DNN and BiLSTM baselines, three task settings are evaluated: single-task recall severity prediction (Y1), single-task root-cause category prediction (Y2), and multi-task prediction of Y1 and Y2.

\noindent \textbf{Training protocol:} We adopted a stratified random 80\%--20\% train--test split based on the primary target variable, i.e., recall class, to preserve the recall severity distribution.  For classical ML and boosting-based models, 10-fold cross-validation was performed on the training set, and the validation trends were found to be consistent with the final test results.
For DNN and BiLSTM+Attention models, a 70\%--10\%--20\% train--validation--test split was used due to the computational cost of repeated cross-validation. The validation set was used for early stopping and learning-rate reduction through ReduceLROnPlateau. For PubMedBERT-based models, the 80\%--20\% train--test split was used, and 
PubMedBERT was fine-tuned jointly with the downstream layers. All experiments were conducted using a fixed random seed to ensure reproducibility of the train--test split. 
For the chronological evaluation, records were split by the event date initiated field: records dated before May 1, 2021 formed the training set (n = 43,403), and those dated on or after that date formed the test set (n = 10,762). We evaluated LightGBM, single-task PubMedBERT, and RecallRisk-BERT on this split, using the same inputs and hyperparameters as in the random-split setting.

\noindent \textbf{Evaluation metrics:} Model performance was evaluated using overall accuracy, macro-averaged precision, recall, and F1-score, and macro-averaged one-vs-rest ROC-AUC. 
We used 10,000 paired bootstrap resamples of the random test set to estimate 95\% confidence intervals for differences in macro-F1, Class~I recall, and Class~I F1. In each replicate, we computed both models' scores on the same resampled test records and recorded their difference. The results are reported in Supplementary Table~S2. Confidence intervals were computed for the random split only; they reflect test-set sampling variability and not variability across training runs.

\noindent \textbf{Parameters:}
DNN and BiLSTM+Attention models were trained using the Adam optimizer with a learning rate of $5 \times 10^{-4}$ and a batch size of 64. These models were trained for a maximum of 50 epochs, with early stopping based on validation loss.
For PubMedBERT-based models, PubMedBERT was fine-tuned jointly with the downstream layers using the AdamW optimizer with a learning rate of $2\times10^{-5}$, a batch size of 128, and 10 epochs. The final-epoch model was evaluated, and the test set was not used for model
or epoch selection.

Class-weighted cross-entropy was used for both tasks. The standard RecallRisk-BERT configuration used the class weights listed in Table~3, with the inverse-frequency weight for Software halved and set to 8.0 to moderate its contribution to the root-cause loss. Equal task coefficients ($\lambda_1=\lambda_2=1$) were used. Class weights controlled the relative importance
of classes within each task, while the task coefficients controlled the contributions of the two losses. Additional experiments with unequal task coefficients ($\lambda_1 \neq \lambda_2$) did not improve performance over the equal-weight configuration.


\begin{table}[!t]
\centering
\caption{Class weights used in the standard RecallRisk-BERT configuration. The Software weight was reduced from its inverse-frequency value and set to 8.0.}
\vspace{0.2cm}
\label{tab:class_weights}
\begin{tabular}{l l c }
\hline
\textbf{Target variable} & \textbf{Class} & \textbf{Weight} \\
\hline
\multirow{3}{*}{\textit{recall class}} & Class I & 4.20 \\
 & Class II & 0.38 \\
 & Class III & 7.30 \\
\hline
\multirow{9}{*}{\textit{root-cause category}} & Unknown & 0.39 \\
 & Design & 0.41 \\
 & Manufacturing & 0.79 \\
 & Material & 0.83 \\
 & Packaging & 1.67 \\
 & Labeling & 2.42 \\
 & Human & 4.42 \\
 & Regulatory & 4.97 \\
 & Software & 8.00  \\
\hline
\end{tabular}
\end{table}

\section{Results}
\label{sec:results}

\subsection{Recall Severity Prediction Performance}
   
\begin{table}[!t]
\caption{Performance comparison of machine learning, deep learning, transformer-based, and multi-task models for recall severity prediction. Recall, precision, and F1-score are reported as macro-averaged metrics to account for class imbalance. The best value for each metric is shown in bold.} \vspace{0.3cm}
\scriptsize
\setlength{\tabcolsep}{5.5pt}
\begin{tabular}{c | c c c ccccc}
\textbf{Input  }                          & \multicolumn{2}{c}{\textbf{Model}}   & \textbf{Accuracy}    & \textbf{Recall}      & 
\textbf{Precision}   & \textbf{F1}          & \textbf{ROC-AUC}     \\
\hline \addlinespace[3pt]
\multirow{4}{*}{\rotatebox{90}{\shortstack{TF-IDF + \\ Tabular}}}& Classical                          & Logistic                           & 0.895      & 0.841      & 0.678      & 0.739      & 0.948      \\
                                  & ML                                 & Regression                         & \multicolumn{1}{l}{} & \multicolumn{1}{l}{} & \multicolumn{1}{l}{} & \multicolumn{1}{l}{} & \multicolumn{1}{l}{} \\
                                  & Classical                          & Linear                             & 0.945      & 0.822     & 0.811      & 0.816      & 0.937      \\
                                  & ML                                 & SVM                                & \multicolumn{1}{l}{} & \multicolumn{1}{l}{} & \multicolumn{1}{l}{} & \multicolumn{1}{l}{} & \multicolumn{1}{l}{} \\
                                    \hline  \addlinespace[2pt]
\multirow{10}{*}{\rotatebox{90}{\shortstack{GloVe + \\ Tabular}}} &  \multirow{2}{*}[3pt]{Boosting}          & \multirow{2}{*}[3pt]{LightGBM}          & \textbf{0.963}      & 0.794      & 0.956     & \textbf{0.856}      & 0.974      \\
                                  &                                    &                                    & \multicolumn{1}{l}{} & \multicolumn{1}{l}{} & \multicolumn{1}{l}{} & \multicolumn{1}{l}{} & \multicolumn{1}{l}{} \\
                                  & \multirow{2}{*}[3pt]{Boosting}          & \multirow{2}{*}[2pt]{XGBoost}           & 0.960      & 0.793      & 0.938     & 0.850      & 0.967      \\
                                  &                                    &                                    & \multicolumn{1}{l}{} & \multicolumn{1}{l}{} & \multicolumn{1}{l}{} & \multicolumn{1}{l}{} & \multicolumn{1}{l}{} \\

                                & \multirow{2}{*}[3pt]{Tree-based}          & \multirow{2}{*}[2pt]{Random Forest}           & 0.949      & 0.694      &  \textbf{0.982}     & 0.779      & \textbf{0.975}      \\
                                  &                                    &                                    & \multicolumn{1}{l}{} & \multicolumn{1}{l}{} & \multicolumn{1}{l}{} & \multicolumn{1}{l}{} & \multicolumn{1}{l}{} \\

                                 & Deep                               & BiLSTM +                           & 0.933      & 0.854      & 0.772      & 0.807      & 0.954      \\
                                  & learning                           & Attention                          & \multicolumn{1}{l}{} & \multicolumn{1}{l}{} & \multicolumn{1}{l}{} & \multicolumn{1}{l}{} & \multicolumn{1}{l}{} \\
                                  & Deep                               & \multirow{2}{*}{DNN}               & 0.928      & 0.833     & 0.756      & 0.791      & 0.948      \\
                                  & learning                           &                                    & \multicolumn{1}{l}{} & \multicolumn{1}{l}{} & \multicolumn{1}{l}{} & \multicolumn{1}{l}{} & \multicolumn{1}{l}{} \\ 
                                  \hline  \addlinespace[3pt]
\multirow{2}{*}{\rotatebox{90}{\shortstack{Text + \\ Tabular}}}      & Transformer                        & PubMedBERT                         & 0.929      & \textbf{0.872}      & 0.765      & 0.806      & 0.962     \\ [3pt] 
                                  & Proposed                           & RecallRisk-BERT                    & 0.953      & 0.865      & 0.840      & 0.852      & 0.961     
\end{tabular}
\label{tab:mainY1}
\end{table}

Table~\ref{tab:mainY1} presents a comparative evaluation of classical machine learning, deep learning, transformer-based, and multi-task learning approaches for recall severity prediction on the random test set. LightGBM with GloVe and tabular features achieved the highest accuracy (0.963) and macro-F1 (0.856), whereas Random Forest obtained the highest macro precision (0.982) and ROC-AUC (0.975). However, Random Forest had lower macro recall (0.694) and macro-F1 (0.779) than LightGBM.

Single-task PubMedBERT achieved the highest macro recall (0.872), followed by RecallRisk-BERT (0.865). The proposed RecallRisk-BERT model achieved an accuracy of 0.953 and a macro-F1 of 0.852 while also predicting root-cause categories within the same framework. 

\subsection{Deep Learning and Multi-task Model Comparison}

\begin{table}[!t]
\caption{Performance comparison of deep learning, transformer-based, and multi-task models for recall severity and root-cause prediction. Y1 denotes recall severity prediction, while Y2 denotes root-cause category prediction. F1-score and recall are reported as macro-averaged metrics. 
} \vspace{0.3cm}
\small
\begin{tabular}{l| llccc}
                                  & \multicolumn{1}{c}{\textbf{Model}} & \multicolumn{1}{c}{\textbf{Task setting}} & \textbf{Y1 - F1}                                & \textbf{Y1 - Recall}                            & \textbf{Y2 - F1}                                \\
                                  \hline
\multirow{10}{*}{\rotatebox{90}{\shortstack{GloVe+ \\ Tabular}}} 

& BiLSTM +                           & \multirow{2}{*}{Single-task Y1}           & \multirow{2}{*}{0.807}                          & \multirow{2}{*}{0.854}                 & \multicolumn{1}{c}{\multirow{2}{*}{\textbf{—}}} \\
                                  & Attention                          &                                           &                                                 &                                                 & \multicolumn{1}{l}{}                            \\
                                  & BiLSTM +                           & \multirow{2}{*}{Single-task Y2}           & \multicolumn{1}{c}{\multirow{2}{*}{\textbf{—}}} & \multicolumn{1}{c}{\multirow{2}{*}{\textbf{—}}} & \multirow{2}{*}{0.657}                          \\
                                  & Attention                          &                                           & \multicolumn{1}{l}{}                            & \multicolumn{1}{l}{}                            &                                                 \\
                                  & DNN                                & Single-task Y1                            & 0.791                                           & 0.833                                           & \multicolumn{1}{c}{\textbf{—}}                  \\
                                  & DNN                                & Single-task Y2                            & \multicolumn{1}{c}{\textbf{—}}                  & \multicolumn{1}{c}{\textbf{—}}                  & 0.661                                           \\
                                  \cline{2-6} \addlinespace[3pt]
                                  & BiLSTM +                           & Multi Task                                & \multirow{2}{*}{0.754}                          & \multirow{2}{*}{0.851}                          & \multirow{2}{*}{0.688}                          \\
                                  & Attention                          & (Y1 + Y2)                                 &                                                 &                                                 &                                                 \\  [3pt]
                                  & \multirow{2}{*}{DNN}               & Multi Task                                & \multirow{2}{*}{0.763}                          & \multirow{2}{*}{0.842}                          & \multirow{2}{*}{0.671}                          \\
                                  &                                    & (Y1 + Y2)                                 &                                                 &                                                 &                                                 \\
                                  \cline{2-6} \addlinespace[3pt]
\multirow{6}{*}{\rotatebox{90}{\shortstack{Text + \\ Tabular}}}   & PubMedBERT                         & Single-task Y1                            & 0.806                                  & \textbf{0.872}                                           & \multicolumn{1}{c}{\textbf{—}}                  \\ [3pt]
                                  & PubMedBERT                         & Single-task Y2                            & \multicolumn{1}{c}{\textbf{—}}                  & \multicolumn{1}{c}{\textbf{—}}                  & \textbf{0.726}                                  \\ [3pt]
                                  & \multirow{2}{*}{RecallRisk-BERT}   & Multi Task                                & \multirow{2}{*}{\textbf{0.852}}                          & \multirow{2}{*}{0.865}                          & \multirow{2}{*}{0.710}                          \\
                                  &                                    & (Y1 + Y2)                                 &                                                 &                                                 &     \\
\end{tabular}
\label{tab:dl_mtl_comparison}
\end{table}

Table~\ref{tab:dl_mtl_comparison} compares recurrent, feed-forward, and transformer-based models for recall severity prediction (Y1) and root-cause category prediction (Y2) in single-task and multi-task settings on the random test set. Compared with single-task PubMedBERT, RecallRisk-BERT achieved higher severity macro-F1 (0.852 vs. 0.806), whereas root-cause macro-F1 decreased from 0.726 to 0.710. The configurations used the same text and tabular inputs and the same class weights for each task, with comparable training conditions.
The paired 95\% bootstrap confidence intervals for both macro-F1 differences excluded zero. In the comparison with LightGBM, the confidence interval for the severity macro-F1 difference included zero, whereas the confidence interval for the root-cause macro-F1 difference favored LightGBM (Supplementary Table S2).

Among the multi-task models, RecallRisk-BERT achieved the highest macro-F1 on both tasks, outperforming the BiLSTM with attention and the DNN. Because these models differ in both text representation and architecture, the comparison does not isolate the contribution of either component.

Supplementary Table~S3 reports precision, recall, F1, and support for all nine root-cause categories on the random test set. Supplementary Figure~S1 shows the corresponding confusion matrix for RecallRisk-BERT. Software and Human had the lowest F1 scores (0.416
and 0.462, respectively). Of the 70 Software records, 15 were misclassified as Design and 10 as Unknown. Of the 269 Human records, 62 were misclassified as Material and 50 as Unknown.

\subsection{Effect of Text Representation}

\begin{table}[!t]
\caption{Effect of text representation and tabular features on recall severity prediction. F1-score and recall are reported as macro-averaged metrics. The best value for each metric is shown in bold.}
\begin{tabular}{l l c c c}
\multicolumn{1}{c}{\textbf{Model}} & \textbf{Text}           & \multicolumn{1}{c}{\textbf{Tabular}} & \multicolumn{1}{c}{\textbf{F1}}     & \multicolumn{1}{r}{\textbf{Recall}} \\
                                   & \textbf{Representation} & \textbf{Features}                    &                                     &                                     \\
                                   \hline
Logistic Regression                & TF-IDF                  & No                                   & \multicolumn{1}{c}{0.717} & \multicolumn{1}{c}{\textbf{0.846}} \\
Logistic Regression                & TF-IDF                  & Yes                                  & \multicolumn{1}{c}{0.739} & \multicolumn{1}{c}{0.841} \\
Linear SVM                         & TF-IDF                  & No                                   & 0.801                     & 0.814                    \\
Linear SVM                         & TF-IDF                  & Yes                                  & 0.816                     & 0.822                     \\
\hline
LightGBM                           & GloVe                   & No                                   & 0.847                    & 0.775                    \\
\multirow{2}{*}{LightGBM}          & GloVe                   & \multirow{2}{*}{No}                  & \multirow{2}{*}{0.844}    & \multirow{2}{*}{0.780}    \\
                                   & (\textit{reason for recall} only) &                                      &                                     &                                     \\
LightGBM                           & GloVe                   & Yes                                  & \multicolumn{1}{c}{\textbf{0.856}} & 0.794                     \\
\hline
LightGBM                           & SentenceBERT            & Yes                                  & 0.820                     & 0.764                    \\
LightGBM                           & PubMedBERT              & Yes                                  & 0.800                     & 0.740    \\
LightGBM                           & BioBERT              & Yes                                  & 0.789                     & 0.730   
\end{tabular}
\label{tab:text_representation}
\end{table}

Table~\ref{tab:text_representation} examines the effect of different text representations and tabular feature integration on recall severity prediction. Among the classical linear models, adding tabular features improved F1-score for both logistic regression and linear SVM, increasing F1 from 0.717 to 0.739, and from 0.801 to 0.816, respectively. However, logistic regression without tabular features achieved the highest recall among the TF-IDF-based models.

Among the LightGBM models, the GloVe representation combined with tabular features achieved the best overall F1-score (0.856), indicating that the integration of dense word embeddings with structured device/regulatory information is beneficial for recall severity classification. The comparison between GloVe-only and GloVe+tabular settings also shows that tabular features improve F1-score, although recall remains lower than that of TF-IDF-based logistic regression.

Interestingly, transformer-derived sentence embeddings such as SentenceBERT, PubMedBERT, and BioBERT did not outperform GloVe-based LightGBM representations in this setting. 
Overall, the results suggest that both representation choice and feature integration substantially affect recall severity prediction performance, with GloVe+tabular features providing the strongest F1-score in the LightGBM setting.

\subsection{Chronological Evaluation}
\label{sec:5.4}

Table~\ref{tab:timesplit} compares the three models under random and chronological evaluation. Macro-F1 was lower for all three models on both tasks under the chronological split. LightGBM ranked first on both tasks under the random split, whereas single-task PubMedBERT led under
the chronological split, with macro-F1 scores of 0.494 for severity and 0.292 for root cause. RecallRisk-BERT ranked second on both tasks (0.458 and 0.278), ahead of LightGBM (0.358 and 0.184). Its severity advantage over single-task PubMedBERT on the random split did not persist on later records. Because the two splits use different test sets, the changes in scores between them are descriptive.

\begin{table}[!t]
\footnotesize
\renewcommand{\arraystretch}{1.3}
\caption{Macro-F1 scores for recall severity and root-cause
prediction under random and chronological splits.}
\begin{tabular}{ l | c  c  c  c}
\multicolumn{1}{c | }{\multirow{2}{*}{\textbf{Model}}} & \textbf{Severity F1}       & \textbf{Severity F1}         & \textbf{Root-cause F1}     & \textbf{Root-cause F1}       \\
 \multicolumn{1}{c}{}                                & \multicolumn{1}{| c  }{Random} & \multicolumn{1}{c}{Chronological} & \multicolumn{1}{c}{Random} & \multicolumn{1}{c}{Chronological}  \\
\hline
LightGBM  (GloVe)                         & \textbf{0.856}   & 0.358                         & \textbf{0.798}                          & 0.184                           \\
PubMedBERT                                            & \multirow{2}{*}{0.806}     & \multirow{2}{*}{\textbf{0.494}}       & \multirow{2}{*}{0.726}     & \multirow{2}{*}{\textbf{0.292}}       \\
(Single-task)                                         &                            &                              &                            &                              \\
RecallRisk-BERT                    & 0.852                       & 0.458                         & 0.710                          & 0.278                          
\end{tabular}
\par\smallskip
\begin{minipage}{\linewidth}
\footnotesize
\textit{Note:} All models also incorporate tabular inputs.
F1 scores are macro-averaged. For single-task PubMedBERT,
the results for the two tasks come from separately
trained models.
\end{minipage}
\label{tab:timesplit}
\end{table}

\subsection{Class I Performance and Error Analysis}

\begin{table}[!t]
\small
\centering
\renewcommand{\arraystretch}{1.2}
\caption{Class I performance on the random test set.}

\begin{tabular}{@{}lrrr@{}}
\hline
\textbf{Model} & \textbf{Precision(\%)} & \textbf{Recall(\%)} & \textbf{F1(\%)} \\
\hline
Random Forest          & \textbf{100.00} & 72.88         & 84.31         \\
LightGBM               & 97.97        & 84.28         & \textbf{90.61} \\
Single-task PubMedBERT & 84.55        & \textbf{90.45} & 87.40        \\
RecallRisk-BERT        & 89.66        & 89.87        & 89.77         \\
\hline
\end{tabular}
\label{tab:class_i_performance}
\par\smallskip
\textit{Note:} The test set contains 859 Class I records.
\end{table}

Table~\ref{tab:class_i_performance} presents the Class I performance of the models on the random test set. Random Forest produced no false positive predictions for Class I (100\% precision), but captured only 72.88\% of the actual Class I records. LightGBM achieved the highest F1 score (90.61\%), whereas the single-task PubMedBERT achieved the highest recall (90.45\%).  RecallRisk-BERT's recall (89.87\%) was higher than LightGBM's and slightly lower than that of single-task PubMedBERT. In terms of precision and F1 score, RecallRisk-BERT ranks below LightGBM but above the single-task PubMedBERT.
RecallRisk-BERT's Class~I recall was 5.59 percentage points higher than LightGBM's (95\% CI: 3.89 to 7.33 percentage points). The confidence intervals for the recall difference with single-task PubMedBERT and the F1 difference with LightGBM both included zero
(Supplementary Table~S2). In contrast, RecallRisk-BERT's Class I F1 was 2.37 percentage points higher than that of single-task PubMedBERT (95\% CI: 0.93 to 3.81).

The confusion matrix for RecallRisk-BERT shows that 772 of the 859 true Class~I records were correctly classified. The model assigned the remaining 87 records (10.13\%) to lower severity classes: 83 to Class~II and 4 to Class~III. All 89 false positives for Class~I were records labeled Class~II.

\subsection{Root-Cause Severity Analysis}
\label{sec:5.6}

Table~\ref{tab:statanalysis} compares mean observed and predicted severity by reported root-cause category on the random test set ($n=10{,}833$). Recall class and reported root-cause category showed a weak
association (Cram\'er's $V=0.121$). Design and Manufacturing had the highest mean observed severity, whereas Labeling and Software had the lowest. The predicted means yielded the same ordering except that Regulatory and Unknown exchanged ranks (Spearman's $\rho$=0.983). This correlation summarizes agreement across nine category means.

\begin{table}[!t]
\footnotesize
\caption{Observed and model-predicted recall severity by root-cause category. Class percentages are calculated within each category. Ground-truth severity scores use Class I = 3, Class II = 2, and Class III = 1. Predicted severity scores are calculated using Equation~\ref{eq:expectedRisk}.}
\begin{tabular}{l c c c c c}
\multicolumn{1}{l}{\textbf{Root-Cause}}  & \textbf{Class I}                  & \textbf{Class II}                 & \textbf{Class III}                & \textbf{Mean GT}                    & \textbf{Mean Predicted}           \\
\textbf{Category}                                         & \multicolumn{1}{c}{\textbf{(\%)}} & \multicolumn{1}{c}{\textbf{(\%)}} & \multicolumn{1}{c}{\textbf{(\%)}} & \multicolumn{1}{c}{\textbf{Severity Score }} & \multicolumn{1}{c}{\textbf{Severity Score}} \\
\hline
Design                                                                  & \textbf{10.96}                    & 86.95                    & 2.09                     & 2.089                       & 2.077                             \\
Human                                                                  & 2.97                              & 91.45                             & 5.58                              & 1.974                                 & 1.989                             \\
Labeling                                                                & 1.78                              & 85.21                             & \textbf{13.02}                             & 1.888                                 & 1.917                             \\
Manufacturing                                                           & 9.64                              & 87.73                             & 2.63                              & 2.070                                 & 2.065                             \\
Material                                                                & 8.60                              & 85.83                             & 5.57                              & 2.030                                 & 2.038                            \\
Packaging                                                              & 5.26                              & 93.31                             & 1.42                              & 2.038                                 & 2.044                             \\
Regulatory                                                              & 2.02                              & \textbf{97.98}                             & 0.00                              & 2.020                                & 2.006                             \\
Software                                                                & 1.43                              & 90.00                             & 8.57                              & 1.929                                 & 1.961                             \\
Unknown                                                                 & 6.51                              & 86.42                             & 7.06                              & 1.995                                 & 2.018                            
\end{tabular}
\label{tab:statanalysis}
\end{table}


\section{Discussion}
\label{sec:discussion}

\subsection{Principal findings}
On the random split, LightGBM achieved the highest macro-F1 on both
tasks, although the paired confidence interval for its difference
from RecallRisk-BERT in severity macro-F1 included zero.
RecallRisk-BERT had higher Class~I recall than LightGBM (89.9\% vs.\
84.3\%; difference, 5.6 percentage points; 95\% CI, 3.9--7.3), with
lower precision (89.7\% vs.\ 98.0\%). Compared with single-task
PubMedBERT, RecallRisk-BERT achieved higher severity macro-F1 but
lower root-cause macro-F1. The paired confidence intervals for both
differences excluded zero. Joint learning also increased Class I F1 relative to single-task PubMedBERT (difference, 2.4 percentage points; 95\% CI, 0.9–3.8). The lower root-cause
performance may reflect the weak association between the two outcomes
(Cramér's V = 0.121).

\subsection{Random versus chronological evaluation}
Macro-F1 declined on both tasks for all three models under
chronological evaluation, with the largest declines for LightGBM.
Single-task PubMedBERT ranked first on both tasks, followed by
RecallRisk-BERT and LightGBM. The severity advantage of joint
training on the random split did not persist on later records.
Possible overlap of recall events in the random split and changes
in record composition over time may both contribute to the gap.
Our comparison does not separate these effects. The larger decline of LightGBM is consistent with greater reliance on recurring product codes, regulation numbers, and specific wording, whereas the fine-tuned transformer models may capture information in recall
narratives that transfers better to later records.

\subsection{Comparison with previous work}
Slivinskis et al.~\cite{Barbosa2025} and the 510(k)-based
studies~\cite{everhart2023,Zhalechian2024,zhu2025} examined recall
occurrence, while Hu et al.~\cite{hu2026} studied recall initiators.
We classify severity and reported root cause in existing recall
records. These differences in prediction targets limit direct
comparison of performance scores. In contrast, the present study covers the full openFDA recall database across all device categories and reports chronological evaluation and confidence intervals.

\subsection{Root-cause categories and recall severity}
The agreement between observed and predicted severity rankings
(Table~9) describes patterns across category means and does not
establish record-level predictive validity. Predictions of Material
or Unknown accounted for 64.8\% of root-cause errors
(Supplementary Figure~S1). Of the 29 misclassified Software records,
15 were predicted as Design. In our taxonomy, software design causes
were assigned to Design and excluded from Software.

\subsection{Limitations and future work}
We used openFDA records as downloaded and could not verify which
inputs were available before FDA classification. The root-cause
findings depend on our grouping of 43 labels into nine categories.
The bootstrap intervals reflect resampling of the random test set
and do not capture variation across training runs. We did not assess the calibration of the probabilities used in the Expected Recall Severity Score, validate the models externally on independent data \cite{liuzzi2026}, or evaluate them in an expert review workflow.

Comparing record-level and event-grouped random splits could help
assess the influence of event overlap. Before considering triage
use, future studies should evaluate Class~I recall at a fixed review
workload using records as first reported.

\section{Conclusion}
\label{sec:conc}

We evaluated RecallRisk-BERT for joint classification of recall
severity and reported root cause in 54,165 product-level FDA recall
records. On the random split, it achieved higher severity macro-F1
than single-task PubMedBERT (0.852 vs.\ 0.806), but lower root-cause
macro-F1 (0.710 vs.\ 0.726). Compared with LightGBM, RecallRisk-BERT
identified a larger proportion of Class~I records (89.9\% vs.\ 84.3\%),
with lower precision (89.7\% vs.\ 98.0\%). On the chronological test
set, all three models scored lower, and single-task PubMedBERT ranked first on both tasks, whereas LightGBM, which ranked first on the random split, ranked last.

The change in model ranking shows the value of testing on later
records. The Class~I results also show why recall and precision
should be reported alongside macro-F1. Before considering triage use,
future evaluations should test the models on reports as first submitted
and measure Class~I recall at a fixed review workload.

\section*{Declaration of generative AI and AI-assisted technologies in the writing process}

During the preparation of this work the authors used ChatGPT (OpenAI) in order to improve the language of the manuscript and assist in revising the text of the manuscript. After using this tool, the authors reviewed and edited the content as needed and take full responsibility for the content of the publication.

\section*{Data availability}
The FDA medical device recall data used in this study are publicly available
from openFDA (https://open.fda.gov/apis/device/recall/download/), accessed
December 2025. The code for data processing, model training, and evaluation will be made publicly available upon acceptance.


\bibliographystyle{elsarticle-num}
\bibliography{refs}

\end{document}